\documentclass{article}
\usepackage{times}
\usepackage{proceedings}
\usepackage{graphicx}
\usepackage{latexsym}
\usepackage{amssymb}

\newtheorem{theorem}{\bf Theorem}
\newcommand{\text}[1]{\mbox{\rm #1}}

\newtheorem{proposition}{\bf Proposition}

\newtheorem{definition}{\bf Definition}
\newtheorem{example}{\bf Example}
\newcommand{\qed}{\rule{2mm}{2mm}}

\newcommand{\M}[1]{\ensuremath{\mathcal{#1}}}

\newcommand{\BD}{{\bf BD}}
\newcommand{\WBD}{{\bf WBD}}
\newcommand{\GBD}{{\bf GBD}}
\newcommand{\BN}{{\bf BN}}
\newcommand{\PL}{{\bf PL}}

\title{Some logics of belief and disbelief}
\author{ {\bf Samir Chopra }
\\schopra@cse.unsw.edu.au\\
Knowledge Systems Group\\
School of Computer Science and Engineering\\
University of New South Wales, Australia\\
\And
{\bf Johannes Heidema}  \\
heidej@unisa.ac.za\\
Department of Mathematics\\
University of South Africa\\
Unisa 0003, South Africa\\
\And
{\bf Thomas Meyer}\\tmeyer@cs.up.ac.za\\Department of Computer Science \\University of Pretoria\\Pretoria 0002, South Africa }

\begin{document}

\maketitle

%

\bibliographystyle{ecai2002}

\begin{abstract}
The introduction of explicit notions of rejection, or disbelief, into logics for knowledge
representation can be justified in a number of ways. Motivations range from the need for
versions of negation weaker than classical negation, to the explicit recording of
classic belief contraction operations in the area of belief change, and the additional
levels of expressivity obtained from an extended version of belief change which includes
disbelief contraction. In this paper we
present four logics of disbelief which address some or all of these intuitions. Soundness and
completeness results are supplied and the logics are compared with respect to
applicability and utility.
\end{abstract}

\section{Introduction}
The introduction of explicit notions of \emph{belief rejection} or \emph{disbelief} into logics
for knowledge representation can be justified in a number of ways. One such justification is found
in the research area of \emph{belief change} \cite{Alchourron-ea:85a,Gardenfors:88a}.
Classic belief change recognises two basic operations on the
beliefs of an agent: \emph{revision}, where the aim is to incorporate a new belief
into an agent's belief state while still maintaining consistency, and \emph{contraction},
where an agent needs to discard one of its beliefs while, at the same time, trying to retain as
many of its remaining beliefs as possible. The argument for introducing explicit disbeliefs into the object
language of choice is based on the conviction that belief contraction should be seen as a
belief change operation on par with revision, and not just as an intermediate step for performing revision,
as it is sometimes presented. The following example, based on an example in \cite{Meyer-ea:2001c},
but originally found
in \cite{Ghose-ea:98a}, illustrates the point.\footnote{%
Observe that this is an example of \emph{base change} \cite{Fuhrmann:91a,Hansson:93c,Rott:96a}.}
\begin{example}
Consider the situation where a totally ignorant agent revises its current (empty) belief base
with $p\rightarrow q$, then contracts with
$q$, and then revises with $p$. In a framework which caters for the explicit representation of beliefs
only, the principle of informational economy\footnote{The methodological injunction that the revision process preserve as many of the old beliefs as possible.}  dictates that there is only one feasible result for this
sequence of base change operations: the set $\{p\rightarrow q, p\}$. The argument against the claim
that this is the only acceptable outcome is that, because
the representational framework does not allow for the explicit recording of the contraction of $q$,
the choice of permissible outcomes is skewed in favour of revision operations.
Contrast this with a situation in which disbelieving
a sentence $\phi$ can be represented in the object language as a sentence of the form
$\overline{\phi}$. In this case the sets $\{p\rightarrow q, {\overline{q}}\}$ and
$\{\overline{q}, p\}$ will be acceptable outcomes as well.
\end{example}
In the example above, classical belief contraction is equated with disbelief \emph{revision}.
But the extended version of belief change in which it is also possible to perform
contraction with
\emph{disbeliefs} has no equivalent in classical belief change. The contraction of
disbeliefs provides for a particularly interesting class of operations if
disbeliefs are thought of as \emph{guards} against the introduction of certain beliefs, in the
spirit of default logic \cite{Reiter:80a}.
Disbelief \emph{contraction} corresponds to the removal of such a guard, which  might then trigger
the addition of previously suppressed beliefs into the current belief set of an agent.
An adequate modelling of this type of extended (dis)belief change requires some level of interaction
between beliefs and disbeliefs, something which has not been dealt with adequately in the
literature. It also presupposes a consequence relation which handles disbeliefs
                appropriately, since it allows for contraction with disbeliefs which are \emph{consequences} of
                previously asserted disbeliefs.

                Another motivation for the introduction of disbeliefs is that it allows us to reason about rejected
                beliefs and their consequences \cite{Slupecki-ea:72a,Gomolinska:98a,Gomolinska-ea:2001a}.
                In this scenario the argument is that in many practical situations it makes more sense to adopt
                a ``negative stance'': to reason about how to reject new beliefs on the basis of
                rejecting old ones. Under these circumstances the focus is on the provision of an appropriate
                consequence relation with regard to disbeliefs.

                Yet another motivation for an explicit representation of disbeliefs is that it can be viewed as
                a weaker version of classical negation. The value of such an additional version of
                negation is that it enables us to resolve well-known paradoxes of belief such as the lottery paradox.
                \begin{example}
                \label{Ex:LP}%
                Given a finite number of lottery tickets issued in a particular week, I express doubt about the
                possibility of any individual ticket winning the lottery, although I do believe that
                exactly one of these tickets will win the lottery.
                \end{example}
                The lottery paradox would be resolved by replacing classical negation (I believe the claim
                that ticket number 113 will not win the lottery) with the much weaker notion
                of disbelief (I disbelieve the claim that ticket number 113 will win the lottery).
                This should not be seen as an \emph{ad hoc} response to a  particular paradox of belief. To the
                contrary,
                such a weaker version of negation is prevalent in many practical situations. Here is another
                example.
                \begin{example}
                \label{Ex:MM}%
                I have two very good friends, Agnetha and Bj\"orn, who are the only suspects in the murder of
                my friends Annifrid and Benny. I refuse to believe that Agnetha was able to commit the murder (represented
                as the disbelief $\overline{a}$), and similarly for Bj\"orn (represented as the disbelief $\overline{b}$).
                Yet, the evidence compels me to believe that one of them committed the crime (represented as $a\vee b$).
                \end{example}
                The challenge is to provide a formal definition of disbelief in such a way that
                sentences of the form $\overline{\phi}$ can be regarded
                as a legitimate way of negating $\phi$, while at the same time ensuring the consistency of
                sets like $\{\overline{\phi}, \overline{\psi}, \phi\vee\psi\}$.

                In this paper we present four different logics of belief and disbelief. Two of these logics,
                \WBD\ and \GBD\, have been used implicitly in work on disbelief, but not in the form
                of a \emph{logic} of belief and disbelief. The first one,
                \WBD, has a very weak notion of consequence with regard to disbeliefs. It has been used
                for the explicit recording of belief contraction operations in \cite{Meyer-ea:2001c}
                and \cite{Ghose-ea:98a}. This logic provides an extremely weak link between
                beliefs and disbeliefs. The relation between beliefs and disbeliefs is
                made only via a notion of consistency, and this is done for the sole purpose of
                providing an accurate model of classical belief contraction in terms of disbeliefs.

                The second logic, \GBD, was presented in \cite{Gomolinska:98a,Gomolinska-ea:2001a},
                albeit in a different form,
                and is a formalisation of the idea that it is useful to reason about rejected beliefs and
                their consequences. The notion of consequence, with regard to disbeliefs,
                associated with \GBD\ is stronger than that of \WBD. We argue
                in section \ref{Sec:GBD} that it is \emph{too} strong. On the other hand, the connection between
                beliefs and disbeliefs in \GBD\ is just as weak as in \WBD. We argue that it needs to
                be strengthened.

                The third logic, \BD, is designed with the express intention of obtaining a weaker version of
                classical negation. Its version of consequence with respect to disbeliefs seems to
                be pitched at just the right level. It is weaker than the version provided by \GBD, but stronger than
                that of \WBD. And unlike the cases for \WBD\ and \GBD\, there is a strong and intuitively plausible
                connection between beliefs and disbeliefs. Indeed, one of the consequences of
                the belief in a sentence $\lnot\phi$ is that $\phi$ will be disbelieved (that is, $\overline{\phi}$
                will hold), as befits a view of disbelief as a weaker version of classical negation.
                It is our contention that \BD\ is the most useful of the logics introduced in this
                paper.

                With the introduction of \BN, the fourth and final logic to be discussed, the aim is
                to take matters a step further by obtaining new beliefs from existing disbeliefs. In the
                process of doing so, however, disbelief collapses into classical negation. That is, disbelieving
                $\overline{\phi}$ becomes equivalent to believing $\lnot\phi$, and \BN\ becomes equivalent to
                classical logic in its level of expressivity.

                In section \ref{Sec:Language}
                we present the language on which all of our
                logics are based. In each of sections \ref{Sec:WBD} to \ref{Sec:BN}
                we present and discuss a logic of belief and disbelief.
                Each logic is equipped with a proof-theoretic and a semantic version of consequence.
                It is shown that these logics are all sound and complete.
                In section \ref{Sec:Conclusion} we draw conclusions and point to future research.

                \subsection{Formal preliminaries}
                We assume a classical propositional logic \PL\ with a language $L_{\PL}$ generated from
                a (possibly countably infinite) set of propositional atoms, together with the
                the usual propositional connectives, and with $\bot$ and $\top$ as canonical representatives of the
                set of contradictions and tautologies, respectively. $V$ is the set of
                classical valuations of \PL. For $\phi\in L_{\PL}$, $M(\phi)$ is the set of models of $\phi$.
                Classical deduction for \PL\ is denoted by $\vdash_{\PL}$.
                For any set $X$ we denote the powerset of $X$ (the set of all subsets of $X$)
                by $\M{P}X$.
                Given a language $L$ and a consequence relation $\vdash_{X}$ from $\M{P}L$ to $L$ we let
                $C_{X}(\Gamma)=\{\alpha\mid\Gamma\vdash_{X}\alpha\}$ be the consequence operation associated
                with $\vdash_{X}$. For example, $C_{\PL}$ denotes classical consequence for propositional logic.
                It is well-known that $C_{\PL}$ satisfies the properties of Inclusion: $X\subseteq C(X)$,
                Idempotency: $C(X)=C(C(X))$, Monotonicity: $X\subseteq Y$ implies $C(X)\subseteq C(Y)$, and
                Compactness: if $\phi\in C(X)$ then $\phi\in C(Y)$ for some finite subset $Y$ of $X$.\footnote{%
                Instead of taking \PL\ as our ``basic logic'' we can work with any logic whose
                associated consequence relation satisfies these four properties.}
                An operation $C$ from $\M{P}L$ to $\M{P}L$ is a \emph{Tarskian consequence operation}
                iff it satisfies Inclusion, Idempotency and Monotonicity.

                \section{A language for beliefs and disbeliefs}
                \label{Sec:Language}%
                The language $L$ on which we base all of the logics to be presented is a simple extension
                of $L_{\PL}$. For each potential belief $\phi$, expressed as a sentence of $L_{\PL}$, there is a
                corresponding potential disbelief $\overline{\phi}$.
                \begin{definition}
                $L=L_{B}\cup L_{D}$, where $L_{B}=L_{\PL}$ is the set of potential beliefs and
                $L_{D}= \{\overline{\phi}\mid\ \phi\in L_{P}\}$, the set of potential disbeliefs.
                An \emph{information set} $\Gamma$ is any subset of $L$. The \emph{beliefs} in $\Gamma$ are defined as
                $\Gamma_{B}=L_{B}\cap\Gamma$. The \emph{disbeliefs} in $\Gamma$ are defined as
                $\Gamma_{D}=L_{D}\cap\Gamma$.
                \end{definition}
                We use $\phi$ and $\psi$ to denote potential beliefs and
                $\alpha$ and $\beta$ to denote arbitrary sentences of $L$ (potential beliefs and disbeliefs).
                The language $L$
                is fairly restrictive in the sense that $\stackrel{-}{\ }$ is not viewed as a propositional connective.
                Thus, we cannot construct sentences of the form $p\vee\overline{q}$, for example. The reason for this
                is twofold. Firstly, in many (but not all) motivations for the introduction of disbeliefs, such a
                level of expressivity is simply not necessary. And secondly, this paper should be seen as a first step
                towards a description of logics of belief and disbelief. Our intention is to treat
                $\stackrel{-}{\ }$ as a full-blown ``propositional'' connective in future research.

                \section{The logic \WBD}
                \label{Sec:WBD}%
                The first logic we consider is relatively weak in terms of the consequences of explicit disbeliefs.
                The motivation for the introduction of \WBD\ is primarily for the explicit expression of classical
                belief contraction. It was used as such in \cite{Ghose-ea:98a} and \cite{Meyer-ea:2001c},
                albeit implicitly. A proof theory for \WBD\ is obtained from the following three inference rules.
                \begin{description}
                \item[(B)] If $\Gamma_{B}\vdash_{\PL}\phi$ then $\Gamma\vdash\phi$
                \item[(D$\bot$)] If $\phi\vdash_{\PL}\bot$ then $\Gamma\vdash\overline{\phi}$
                \item[(WD)] If $\overline{\psi}\in\Gamma_{D}$ and $\phi\vdash_{\PL}\psi$
                          then $\Gamma\vdash\overline{\phi}$
                 \end{description}
                (B) is a supraclassicality requirement. (D$\bot$) requires that contradictions
                always be disbelieved, analogous to the case of tautologies always being believed.
                (WD) requires that disbelieving a sentence $\overline{\psi}$ leads to a disbelief
                in all those sentences classically stronger than $\psi$.
                \begin{definition}
                In \WBD, $\alpha$ can be deduced from an information set $\Gamma$, written
                as $\Gamma\vdash_{\mathrm{\WBD}}\alpha$, iff $(\Gamma,\alpha)$ is in the smallest binary relation
                from $\M{P}L$ to $L$ closed under (B), (D$\bot$), and (WD).
                \end{definition}
                The logic \WBD\ has a well-behaved Tarskian consequence operation.
                \begin{proposition}
                $C_{\mathrm{\WBD}}$ satisfies Inclusion, Idempotency, Monotonicity and Compactness.
                \end{proposition}
                Observe that beliefs and disbeliefs are completely decoupled in \WBD. In particular, \WBD\
                satisfies the following two properties which show that disbeliefs and beliefs are only derivable from the set of disbeliefs and beliefs respectively:
                \begin{description}
                \item[(B${\not\rightarrow}$D)] $\forall\Omega\subseteq L_{B}$ and $\phi\in L_{B}$,
                $\Gamma\vdash\overline{\phi}$ iff $\Gamma_{D}\cup\Omega\vdash\overline{\phi}$
                \item[(D${\not\rightarrow}$B)] $\forall\Delta\subseteq L_{D}$ and $\phi\in L_{B}$,
                       $\Gamma\vdash\phi$ iff $\Gamma_{B}\cup\Delta\vdash\phi$
                \end{description}
                \begin{proposition}
                $\vdash_{\mathrm{\WBD}}$ satisfies (B${\not\rightarrow}$D) and (D${\not\rightarrow}$B).
                \end{proposition}
                In fact, the only way in which beliefs and disbeliefs are related in \WBD\ is by way of the
                notion of \emph{\WBD-inconsistency}.
                \begin{definition}
                $\Gamma$ is \emph{WB-inconsistent} iff $\Gamma_{B}\vdash_{\mathrm{\WBD}}\bot$. $\Gamma$  is
                \emph{WD-inconsistent} iff $\Gamma_{D}\vdash_{\mathrm{\WBD}}\overline{\top}$.
                $\Gamma$ is \emph{\WBD-inconsistent} iff $\Gamma\vdash_{\mathrm{\WBD}}\phi$ and
                $\Gamma\vdash_{\mathrm{\WBD}}\overline{\phi}$
                for some $\phi\in L_{B}$.
                \end{definition}
                We have similar versions of inconsistency for the other logics too.

                We now turn to the semantics for \WBD. A useful intuition is to think of the beliefs in $\Gamma$ as
                the information an agent has acquired on its own, and each disbelief $\overline{\phi}$ as
                information it has obtained from a different source, informing it that $\phi$ does \emph{not} hold.
                The agent has more faith in its own capabilities than in those of its sources, and information
                obtained from its sources is therefore seen as less reliable.
                The information obtained  from a specific source is independent
                of the beliefs of the agent
                and the information obtained from other sources. A model for \WBD\ consists of a set of valuations,
                corresponding to the worlds that the agent regards as possible, together with a set of sets
                of valuations, with each element of this set corresponding to the worlds that a particular
                source of the agent regards as \emph{possible}. A potential belief is satisfied in a model if it is true
                in all the worlds that the agent regards as possible, and a potential disbelief $\overline{\phi}$
                is satisfied in a model if at least one of the sources regards $\phi$ as impossible.
                Satisfaction is denoted by $\Vdash$.
                \begin{definition}
                A \emph{\WBD-model} $\M{M}$ is an ordered pair $(M,\M{N})$ where $M\subseteq V$ and
                $\emptyset\subset \M{N}\subseteq\M{P}V$. For $\phi\in L_{B}$, $\M{M}\Vdash\phi$ iff
                $M\subseteq M(\phi)$.
                For $\overline{\phi}\in L_{D}$, $\M{M}\Vdash\overline{\phi}$ iff
                $\exists N\in \M{N}$ such that $N\subseteq M(\lnot\phi)$.
                \end{definition}
                In a \WBD-model $\M{M}=(M,\M{N})$, $M$ represents the models of the beliefs of the agent, and
                each $N\in\M{N}$ represents the models of a sentence that the source associated with
                $N$ holds to be possible. We require an agent to have at least one source
                of information. That is, we require that $\M{N}\neq\emptyset$.
                Entailment for \WBD\ (denoted by $\vDash_{\mathrm{\WBD}}$) is then defined in the normal
                model-theoretic fashion.
                \begin{definition}
                $\M{M}\Vdash\Gamma$ iff $\M{M}\Vdash\alpha\ \forall\alpha\in\Gamma$.
                $\Gamma\vDash_{\WBD}\alpha$ iff $\M{M}\Vdash\alpha$
                for every \WBD-model $\M{M}$ s.t. $\M{M}\Vdash\Gamma$.
                \end{definition}
                It turns out that the logic \WBD\ is sound and complete.
                \begin{theorem}
                \label{Thm:WBD-Rep}%
                $\Gamma\vdash_{\mathrm{\WBD}}\alpha$ iff $\Gamma\vDash_{\mathrm{\WBD}}\alpha$, for all $\alpha\in L$.
                \end{theorem}
                Based on the semantics of \WBD\ we can also define appropriate notions of (un)satisfiability which,
                via theorem \ref{Thm:WBD-Rep}, can be shown to coincide with inconsistency.
                We postpone presentation of these details to the full version of this paper.

                The logic \WBD\ is, essentially, the logic on which the work in \cite{Ghose-ea:98a} and
                \cite{Meyer-ea:2001c} is based. Both these papers argue
                for an explicit expression of disbeliefs in order to maintain a record of belief contraction.
                Although \WBD\ seems adequate for this purpose, at least in simple settings, two types of
                criticisms can be levelled at it: the extreme weakness of the notion of consequence associated with
                disbeliefs, and the complete decoupling of beliefs and disbeliefs. These weaknesses become apparent
                in scenarios where it is appropriate to perform various belief change operations on beliefs
                as well as disbeliefs. So, while contracting with a belief might correspond to revising with a
                disbelief, contracting with a \emph{disbelief} has no equivalent in classical belief change. And such
                an operation is useful if disbeliefs are thought of as \emph{guards} against
                the introduction of certain information, in the spirit of default logic \cite{Reiter:80a}.
                The removal of such a guard might then trigger the
                addition of beliefs into an agent's current belief set. An adequate modelling of
                this type of extended (dis)belief change requires some level of interaction between
                beliefs and disbeliefs. It also presupposes a consequence relation which handles disbeliefs
                appropriately, since it allows for contraction with disbeliefs which are \emph{consequences} of
                previously asserted disbeliefs. In section \ref{Sec:GBD}
                we consider an attempt to rectify the weakness of consequences derived from disbeliefs.
                In section \ref{Sec:BD} we address this issue as well, and we also consider the
                the provision of a link between beliefs and disbeliefs.

                \section{The logic \GBD}
                \label{Sec:GBD}%
                One way in which to address the weakness of the consequences to be derived from disbeliefs is to
                regard consequence, with respect to disbeliefs, as the exact dual of consequence with respect
                to beliefs. This is the approach followed in \cite{Gomolinska-ea:2001a} where, interestingly enough,
                the reason for defining such a logic is to define disbelief change.

                Let $\overline{\Gamma}_{D}=\{\lnot\phi\mid\overline{\phi}\in\Gamma_{D}\}$
                and consider the following rule:
                \begin{description}
                \item[(GD)] If $\overline{\Gamma}_{D}\vdash_{\PL}\lnot\phi$ then
                $\Gamma\vdash\overline{\phi}$
                \end{description}
                (GD) requires that the consequences of disbeliefs be the exact dual of classical consequence.
                The proof theory for the logic \GBD\ is then obtained from \WBD\ by replacing (WD) with (GD).
                \begin{definition}
                In \GBD\, $\alpha$ can be deduced from an information set $\Gamma$, written
                as $\Gamma\vdash_{\mathrm{\GBD}}\alpha$, iff $(\Gamma,\alpha)$ is in the smallest binary relation
                from $\M{P}L$ to $L$ closed under (B), (D$\bot$), and (GD).
                \end{definition}
                \GBD\ also has a well-behaved Tarskian consequence operation.
                \begin{proposition}
                $C_{\mathrm{\GBD}}$ satisfies Inclusion, Idempotency, Monotonicity and Compactness.
                \end{proposition}
                The logic \GBD\ satisfies the following property, which was proposed in \cite{Gomolinska:98a}.
                \begin{description}
                \item[(Rej)]If $\Gamma\vdash\overline{\beta}$ and
                $\Gamma\vdash\overline{\lnot(\alpha\rightarrow\beta)}$ then
                $\Gamma\vdash\overline{\alpha}$
                \end{description}
                (Rej) can be thought of as a version of the inference rule Modus Tollens. In fact, if disbelief is replaced
                with classical negation, (Rej) coincides exactly with Modus Tollens.

                \GBD\ also has a complete decoupling of beliefs and disbeliefs, as the following
                proposition shows.
                \begin{proposition}
                $\vdash_{\mathrm{\GBD}}$ satisfies (B${\not\rightarrow}$D) and (D${\not\rightarrow}$B).
                \end{proposition}
                The three versions of inconsistency for \GBD\ are defined as for \WBD.
                \begin{definition}
                $\Gamma$ is \emph{GB-inconsistent} iff $\Gamma_{B}\vdash_{\mathrm{\GBD}}\bot$.
                $\Gamma$ is \emph{GD-inconsistent} iff $\Gamma_{D}\vdash_{\mathrm{\GBD}}\overline{\top}$.
                $\Gamma$ is \emph{\GBD-inconsistent} iff $\Gamma\vdash_{\mathrm{\GBD}}\phi$ and
                $\Gamma\vdash_{\mathrm{\GBD}}\overline{\phi}$
                for some $\phi\in L_{B}$.
                \end{definition}
                Observe that beliefs and disbeliefs are related only by \GBD-inconsistency.

                A semantics for \GBD\ is obtained by considering only those \WBD-models in which $\M{N}$ has a single element.
                Intuitively, all disbeliefs are obtained from a single source. This ensures that disbeliefs can
                be combined to obtain new disbeliefs, which allows for a stronger notion of consequence regarding
                disbeliefs. \GBD-entailment are defined as for \WBD.
                \begin{definition}
                A \emph{\GBD-model} is an ordered pair $\M{M}=(M,N)$ where $M,N\subseteq V$.
                For $\phi\in L_{B}$, $\M{M}\Vdash\phi$ iff $M\subseteq M(\phi)$.
                For $\overline{\phi}\in L_{D}$, $\M{M}\Vdash\overline{\phi}$ iff
                $N\subseteq M(\lnot\phi)$.
                $\M{M}\Vdash\Gamma$ iff
                $\M{M}\Vdash\alpha\ \forall\alpha\in\Gamma$.
                $\Gamma\vDash_{\GBD}\alpha$ iff $\M{M}\Vdash\alpha$
                for every \GBD-model $\M{M}$ s.t. $\M{M}\Vdash\Gamma$.
                \end{definition}
                The logic \GBD\ is sound and complete.
                \begin{theorem}
                \label{Thm:GBD-Rep}%
                $\Gamma\vdash_{\mathrm{\GBD}}\alpha$ iff $\Gamma\vDash_{\mathrm{\GBD}}\alpha$, for all $\alpha\in L$.
                \end{theorem}
                We postpone a full description of the appropriate versions of unsatisfiability for \GBD, which
                coincide with the three versions of inconsistency for \GBD, to the full version
                of this paper.

                The crucial difference between \WBD\ and \GBD\ is that \GBD\ combines disbeliefs to obtain
                new disbeliefs, which allows for a much stronger notion of consequence with respect to disbeliefs.
                In particular, \GBD\ satisfies the following property.
                \begin{description}
                \item[(D$\vee$)] If $\Gamma\vdash\overline{\phi}$ and $\Gamma\vdash\overline{\psi}$ then
                                 $\Gamma\vdash\overline{\phi\vee\psi}$
                \end{description}
                It is our contention that such a property is undesirable for a notion of disbelief.
                Since disbelief is \emph{not} intended to be equivalent to classical negation, it is reasonable
                to require that it be possible to express notions not expressible in classical logic. One of these
                is \emph{agnosticism}, in which an agent refuses to commit to a potential belief or its negation.
                Formally, this amounts to both $\overline{\phi}$ and $\overline{\lnot\phi}$
                being consequences of a consistent information set $\Gamma$. But if the consequence relation satisfies
                (D$\vee$), it means that the agent is forced to accept $\overline{\phi\vee\lnot\phi}$, and therefore
                $\overline{\top}$ as well. Disbelieving a tautology amounts to GD-inconsistency.
                It is the analogue
                of classical inconsistency (believing the negation of the tautology), but with classical
                negation replaced by disbelief. Now, if agnosticism in the sense described above leads to
                inconsistency with respect to disbeliefs, it is an indication that the consequence relation
                for \GBD\ is too strong to account for a proper treatment of disbeliefs. In section \ref{Sec:BD}
                we consider a logic which seems to be pitched at the right level in this regard.

                \section{The logic \BD}
                \label{Sec:BD}%
                In this section we introduce a logic in which there is interaction between beliefs and disbeliefs.
                Disbelief is seen as a weaker notion of classical negation. As a result, believing the negation
                of a sentence also results in disbelieving that sentence, although the converse relationship
                does not hold. Interestingly enough, this coupling of beliefs and disbeliefs comes about
                as a result of the introduction of the following inference rule,
                which looks like a simple strengthening of the consequence
                relation with respect to disbeliefs.
                \begin{description}
                \item[(D)] If $\overline{\psi}\in\Gamma_{D}$ and $\Gamma_{B}\cup\{\phi\}\vdash_{\PL}\psi$
                         then $\Gamma\vdash\overline{\phi}$
                \end{description}
                (D) requires that disbelieving a sentence $\psi$ also leads to a disbelief
                in those sentences classically stronger than $\psi$, \emph{but with respect to} $\Gamma_{B}$.
                Observe that (WD) is the special case of (D) where $\Gamma_{B}=\emptyset$.
                The difference between the rules (GD), (WD) and (D) is perhaps best brought out through an example:
                \begin{example} Consider our earlier example of the murder mystery. Then  (WD) commits me to the following: If I refuse to believe that Agnetha killed Annifrid and Benny,
                then I also
                refuse to believe that Agnetha and Bj\"orn killed them (a similar commitment is enforced
                by (D) and (GD)). However, in addition,
                (D) commits me to:  If I believe that Agnetha is a Swede, I believe that if you killed
                Annifrid and Benny then you are a murderer, and I refuse to believe that
                a Swede
                can be a murderer, then I also refuse to believe that Agnetha could have
                killed Annifrid and Benny. This argument cannot be made with either (WD)
                or (GD). To further illustrate the difference, consider what
                (GD) requires us to infer in the lottery example. Consider for the time being, a version with
                just two tickets.
                I believe that exactly one of $t_1$
                or $t_2$
                will win the lottery; I refuse to believe that $t_1$ wins the lottery and
                similarly for $t_2$.
                Then (GD) compels me to refuse to believe that exactly one of the two will
                win. Therefore
                I end with a contradiction. In contrast, (D) and (WD) do not require such a commitment.
                \end{example}
                From the example above it should be clear that both (D) and (GD) are stronger
                than (WD). Furthermore, as the second part of the example makes clear, since (D)
                is stronger than (WD), some conclusions obtained from the former will not be obtainable from the latter, but others will. The example provided is one of those obtainable
                from (D) but not (WD). Since (D) and (GD) are incomparable in strength, they will
                have some conclusions that coincide (as above) but there will also be conclusions
                obtained from (D) which are not obtainable from (GD) and vice-versa. The example provided for (D) is an instance of a conclusion obtainable from (D) but not from (GD).
                \begin{definition}
                In \BD, $\alpha$ can be deduced from an information set $\Gamma$, written
                as $\Gamma\vdash_{\mathrm{\BD}}\alpha$, iff $(\Gamma,\alpha)$ is in the smallest binary relation
                from $\M{P}L$ to $L$ closed under (B), (D$\bot$), and (D).
                \end{definition}
                \BD\ has a well-behaved Tarskian consequence operation:
                \begin{proposition}
                $C_{\BD}$ satisfies Inclusion, Idempotency, Monotonicity and Compactness.
                \end{proposition}
                The manner in which disbeliefs are obtained from beliefs in \BD\ can be expressed by the following
                property which links up a belief in $\lnot\phi$ to a disbelief in $\phi$. Observe that neither \WBD\ nor
                \GBD\ satisfies ($B\rightarrow D$).
                \begin{description}
                \item[(B$\rightarrow$D)] If $\Gamma\vdash\lnot\alpha$ then $\Gamma\vdash\overline{\alpha}$
                \end{description}
                \begin{proposition}
                $\vdash_{\mathrm{\BD}}$ satisfies (B$\rightarrow$D) and  does not
                satisfy (B${\not\rightarrow}$D).
                \end{proposition}
                The logic \BD\ does not support any connection between beliefs and disbeliefs in the
                converse direction, though, as the following result shows.
                \begin{proposition}
                $\vdash_{\mathrm{\BD}}$ satisfies (D${\not\rightarrow}$B).
                \end{proposition}
                The different notions of inconsistency for \BD\ are defined in the same way as for \WBD\ and \GBD.
                \begin{definition}
                An information set $\Gamma$ is \emph{B-inconsistent} iff $\Gamma_{B}\vdash_{\mathrm{\BD}}\bot$,
                \emph{D-inconsistent} iff $\Gamma_{D}\vdash_{\mathrm{\BD}}\overline{\top}$,
                and \emph{\BD-inconsistent} iff $\Gamma\vdash_{\mathrm{\BD}}\phi$ and
                $\Gamma\vdash_{\mathrm{\BD}}\overline{\phi}$
                for some $\phi\in L_{B}$.
                \end{definition}
                Because there is interaction between beliefs and disbeliefs in \BD, there is also
                a connection between the different notions of inconsistency for \BD. In particular, we have the following
                results.
                \begin{proposition}
                \label{Prop:BD=D}%
                If $\Gamma$ is B-inconsistent then it is also \BD-inconsistent, but the converse does not hold.
                $\Gamma$ is \BD-inconsistent iff it is D-inconsistent.
                \end{proposition}
                In the logic \BD, then, \BD-inconsistency collapses into D-inconsistency. Given the intuition
                of disbelief as a weaker version of classical negation, this is a particularly
                desirable state of affairs. In classical logic, asserting both $\phi$ and
                $\lnot\phi$ is tantamount to the assertion that $\lnot\top$ is the case, while in \BD,
                asserting both $\phi$ and $\overline{\phi}$ amounts to the assertion that $\overline{\top}$ is the case.

                Observe that \BD-inconsistency (or D-inconsistency) amounts to disbelieving the tautology, which leads
                to a disbelief in every sentence in $L_{B}$. So, while B-inconsistency, like classical consistency,
                leads to the acceptance of \emph{every} sentence in the language, \BD-inconsistency leads only
                to the acceptance of every \emph{disbelief} in the language. \BD-inconsistency can thus be seen as a weaker
                version of classical inconsistency. We regard this as a particularly attractive feature of the
                logic \BD. The logics \WBD\ and \GBD\ also have similar features, but the connection between classical
                inconsistency and the weaker version of inconsistency, based on disbeliefs, is not
                as intuitively appealing.

                The semantics for \BD\ is obtained by considering only those \WBD-models for which
                every $N\in\M{N}$ is a subset of $M$. That is, the worlds that the sources of an agent
                may regard as possible have to be worlds that the agent itself regards as possible.
                \BD-entailment are defined as for \WBD-entailment and \GBD-entailment.
                \begin{definition}
                A \emph{\BD-model} is a tuple $(M,\M{N})$ where $M\subseteq V$ and
                $\emptyset\subset \M{N}\subseteq\M{P}M$.
                For a \BD-model $\M{M}=(M,\M{N})$ and $\phi\in L_{B}$, $\M{M}\Vdash\phi$ iff
                $M\subseteq M(\phi)$.
                For $\overline{\phi}\in L_{D}$, $\M{M}\Vdash\overline{\phi}$ iff
                $\exists N\in \M{N}$ s.t. $N\subseteq M(\lnot\phi)$.
                $\M{M}\Vdash\Gamma$ iff
                $\M{M}\Vdash\alpha\ \forall\alpha\in\Gamma$.
                $\Gamma\vDash_{\BD}\alpha$ iff $\M{M}\Vdash\alpha$
                for every \BD-model $\M{M}$ s.t. $\M{M}\Vdash\Gamma$.
                \end{definition}
                The logic \BD\ is sound and complete.
                \begin{theorem}
                \label{Thm:BD-Rep}%
                $\Gamma\vdash_{\mathrm{\BD}}\alpha$ iff $\Gamma\vDash_{\mathrm{\BD}}\alpha$, for all $\alpha\in L$.
                \end{theorem}
                Based on the semantics of \BD\ we can define appropriate notions of (un)satisfiability which
                can be shown to coincide with inconsistency.
                We postpone presentation of these details to the full version of this paper.

                We conclude this section by pointing out that \BD\ is able to handle examples such as the lottery paradox
                and its variants (cf. examples \ref{Ex:LP} and \ref{Ex:MM})
                in a manner that is intuitively satisfactory. It can be shown that any information set of the
                form $\Gamma=\{\overline{\phi_{1}},\ldots,\overline{\phi_{n}},\bigvee_{i=1}^{i\leq n}\phi_{i}\}$
                is neither B-inconsistent, nor \BD-inconsistent.\footnote{%
                By proposition \ref{Prop:BD=D} this means that $\Gamma$ is also not D-inconsistent.}
                \BD\ thus allows us, for example, to disbelieve the fact that any particular ticket will
                win the lottery, while still believing that exactly one of the tickets will win the lottery,
                without collapsing into \emph{any} kind of inconsistency.

                \section{The logic \BN}
                \label{Sec:BN}%
                We have seen that \BD\ allows for the generation of disbeliefs from beliefs. An interesting
question is whether it makes sense to do the opposite; that is, to generate beliefs from
disbeliefs. We consider two ways of doing so. For the first one, note that the generation
of disbeliefs from beliefs in \BD\ is achieved by the inference rule (D). Now, it is easily
verified that, in the presence of (B) and (D$\bot$), (D) is equivalent to the following
property:
\begin{description}
\item[(D$'$)] If $\Gamma\vdash\overline{\psi}$ and $\Gamma\cup\{\phi\}\vdash\psi$
         then $\Gamma\vdash\overline{\phi}$
\end{description}
On the basis of this equivalence we consider the following property:
\begin{description}
\item[(B$'$)] If $\Gamma\vdash\psi$ and $\Gamma\cup\{\overline{\phi}\}\vdash\overline{\psi}$
         then $\Gamma\vdash\phi$
\end{description}
(B$'$) asserts that if I currently believe $\psi$, and if the addition of $\phi$ as a
disbelief leads me to disbelieve $\psi$, then $\phi$ should be one of my current
beliefs. It is analogous to (D$'$), but with the roles
of beliefs and disbeliefs reversed. It turns out, however, that (B$'$) is a derived
rule of the logic \BD.
\begin{proposition}
$\vdash_{\mathrm{\BD}}$ satisfies the property (B$'$).
\end{proposition}
A more direct way to obtain new beliefs from current disbeliefs is to consider the
converse of the property (B$\rightarrow$D).
\begin{description}
\item[(D$\rightarrow$B)] If $\Gamma\vdash\overline{\phi}$ then $\Gamma\vdash\lnot\phi$.
\end{description}
Observe that \WBD, \GBD\ and \BD\ do not satisfy (D$\rightarrow$B). A proof theory for the logic we call \BN\
is then obtained by adding (D$\rightarrow$B) to the
inference rules of \BD.
\begin{definition}
For $\alpha\in L$ and $\Gamma\subseteq L$, $\alpha$ can be deduced from $\Gamma$ in \BN, written
as $\Gamma\vdash_{\BN}\alpha$, iff $(\Gamma,\alpha)$ is in the smallest binary relation
from $\M{P}L$ to $L$ closed under (B), (WD), (D$\bot$) and (D$\rightarrow$B).
\end{definition}
The introduction of (D$\rightarrow$B) does indeed give us a logic that is stronger than \BD. It turns
out, however, that disbelief now collapses into classical negation
\begin{theorem}
For every $\phi\in L_{B}$, $\Gamma\vdash_{\BN}\overline{\phi}$ iff $\Gamma\vdash_{\BN}\lnot\phi$.
\end{theorem}
\BN\ is therefore exactly as expressive as \PL.

\section{Conclusion and future research}
\label{Sec:Conclusion}%
Of the four logics of belief and disbelief presented in this paper, the logic \BD\ is the most deserving of
such a label. \BD\ covers all the motivations we have mentioned for the explicit introduction of
disbeliefs. Contraction with a belief $\phi$ can be represented explicitly in \BD\ as
revision with the disbelief $\overline{\phi}$, just as it can in \WBD\ and \GBD. This ability has
a useful side-effect as well. In classical belief change, the result of the
pathological case in which a belief set is contracted with a tautology, is taken to be the belief set
itself, primarily because it is unclear what else the result could be.
In fact, it is the
only case in which a contraction does not succeed. But in \BD\ (as in \WBD\ and \GBD), contraction by
$\phi$ corresponds to a revision by $\overline{\phi}$. So contraction by a tautology is
equivalent to revising with the sentence $\overline{\top}$, a disbelief in the tautology.
Furthermore, it can be verified that a disbelief in the tautology results in
\emph{disbelieving} all (propositional) sentences, but has no effect on the
current \emph{beliefs}. The explicit introduction of disbeliefs thus enables us to
devise a result which is much more in line with our intuitions.

Like \WBD\ and \GBD, \BD\ also provides a well-behaved notion of consequence with respect to
disbeliefs. In \BD, consequence with respect to disbeliefs is stronger than in \WBD\ but weaker than in \GBD.
In particular, (D) is satisfied in \BD\ but not in \WBD, and \BD\ does not satisfy (D$\vee$),
a property satisfied by \GBD. As a result of not satisfying (D$\vee$), consequence
for \BD\ is weak enough to allow for the expression of \emph{agnosticism} (disbelieving both
$\phi$ and $\lnot\phi$) without collapsing into inconsistency.
In this respect it is superior to the logic \GBD.
At the same time, there is a link between
beliefs and disbeliefs in \BD\ which is lacking in both \WBD\ and \GBD, and which ensures that
\BD\ is a suitable base logic for an extended version of belief change in which it is
possible to contract with disbeliefs as well. The link between beliefs and disbeliefs
also ensures that disbelief, as defined in \BD, is an appropriate weaker version of classical negation,
as is apparent from the fact that \BD\ satisfies (B$\rightarrow$D) and (D$\not\rightarrow$B), as well
as from the intuitively acceptable manner in which the lottery paradox and its variants are handled.

If disbelief is truly to be viewed as a weaker version of negation, it is necessary to conduct
a proper investigation into the connection between disbelief, classical negation,
and the other propositional connectives. In order to do so, it is necessary to
treat $\stackrel{-}{\ }$ as full-blown propositional connective, in which
sentences such as $p\vee\overline{q}$, $\lnot\overline{p}$, and $\overline{\overline{p}}$
have a well-defined meaning.

There is an obvious connection between beliefs and disbeliefs in \BD\ and the modal
logic operators of necessity and possibility, specifically in the epistemic logic {\bf KD45} \cite{Hughes-ea:72a}. The statement
$\Box\phi$ in {\bf KD45} corresponds to the assertion that $\phi$ is believed, while
$\Diamond\lnot\phi$ corresponds to the assertion that $\lnot\phi$ is possible, and hence that
$\phi$ is disbelieved. Observe, though,
that statements in \BD\ are object-level assertions, or assertions from a first person
perspective (``The sky is blue''), while the corresponding statements in epistemic logics
are meta-level assertions, or assertions from a third person perspective
(``The agent believes the sky is blue''). However, the differences run slightly deeper and are easily illustrated. To compare \BD\ with {\bf KD45} we need to restrict {\bf KD45} to
sentences of the form $\Box \phi$ and $\Diamond \phi$.
Then {\bf KD45} satisfies (B), (D$\bot$) and (D). So it is at least as strong
as
\BD. It does not satisfy  ($B\not\rightarrow D$), and satisfies
($D\not\rightarrow B$).
It does not satisfy (Rej). It supports agnosticism (I disbelieve $\phi$ as
well as $\neg\phi$) but does not satisfy ($D\vee$). In these respects it has the same
behaviour as \BD.
But it is stronger than \BD. This can be seen by looking at what happens
when the
tautology is disbelieved. In {\bf KD45} (indeed, in any normal modal
logic i.e., one containing the axiom schema {\bf K}),
if $\Diamond (\lnot\top)$ can be deduced from $\Gamma$, then so can $\Box \phi$ and $\Diamond \phi$ for every
propositional
sentence $\phi$. That is, disbelieving $\top$ in {\bf KD45} leads to a belief as well as a disbelief in
every propositional sentence. In  contrast disbelieving the tautology in \BD\ leads to a
disbelief in
every propositional sentence, but our beliefs remain unaffected. This is a particularly desirable property---as pointed out in the discussion following Proposition 8.

In the full version of the paper we plan to develop a restricted version of {\bf KD45} as defined above
as a separate logic of belief and disbelief.
Developing an axiomatization will be the primary task since the normal {\bf KD45} characterization requires
a more expressive language than we possess at the moment. Furthermore, we will provide a formal proof to the effect that the semantics of the logic \BD\ is not adequately captured by a class of Kripke models. Applications of the logics that we have developed to belief revision is a non-trivial task that needs  separate study.

\section{Acknowledgements}The first author would like to thank the University of South Africa in Pretoria for the invitation and financial support to visit its Departments of Mathematics and Computer Science in November 2001, when this paper was conceived.
The authors would also like to thank Aditya Ghose for some insightful comments and suggestions.
\bibliography{master}

\end{document}